\documentclass{article}

 \usepackage[main, final]{neurips_2026}

\usepackage[utf8]{inputenc} 
\usepackage[T1]{fontenc}    
\usepackage{hyperref}       
\usepackage{url}            
\usepackage{booktabs}       
\usepackage{amsfonts}       
\usepackage{nicefrac}       
\usepackage{microtype}      
\usepackage{xcolor}         

\usepackage{amsmath}
\usepackage{graphicx}
\usepackage{multirow}
\usepackage{multicol}

\usepackage{algorithm}
\usepackage{algorithmic}

\title{Mesh BDF: Barycentric Dominance Field for 3D Native Mesh Generation}

\author{%
  \textbf{Gaochao Song$^{1,2}$  \quad Haohan Weng$^3$ \quad Luo Zhang$^{4}$ \quad Zibo Zhao$^3$ \quad Shenghua Gao$^{1,2}$} \\ 
  $^1$HKU, $^2$Shenzhen Loop Area Institute,  $^3$Tencent Hunyuan 3D, $^4$NTU \\
  \url{https://gaochao-s.github.io/pages/MeshBDF/}
}

\begin{document}

\maketitle

\begin{figure}[htbp]
    \centering
    \includegraphics[width=0.9\textwidth]{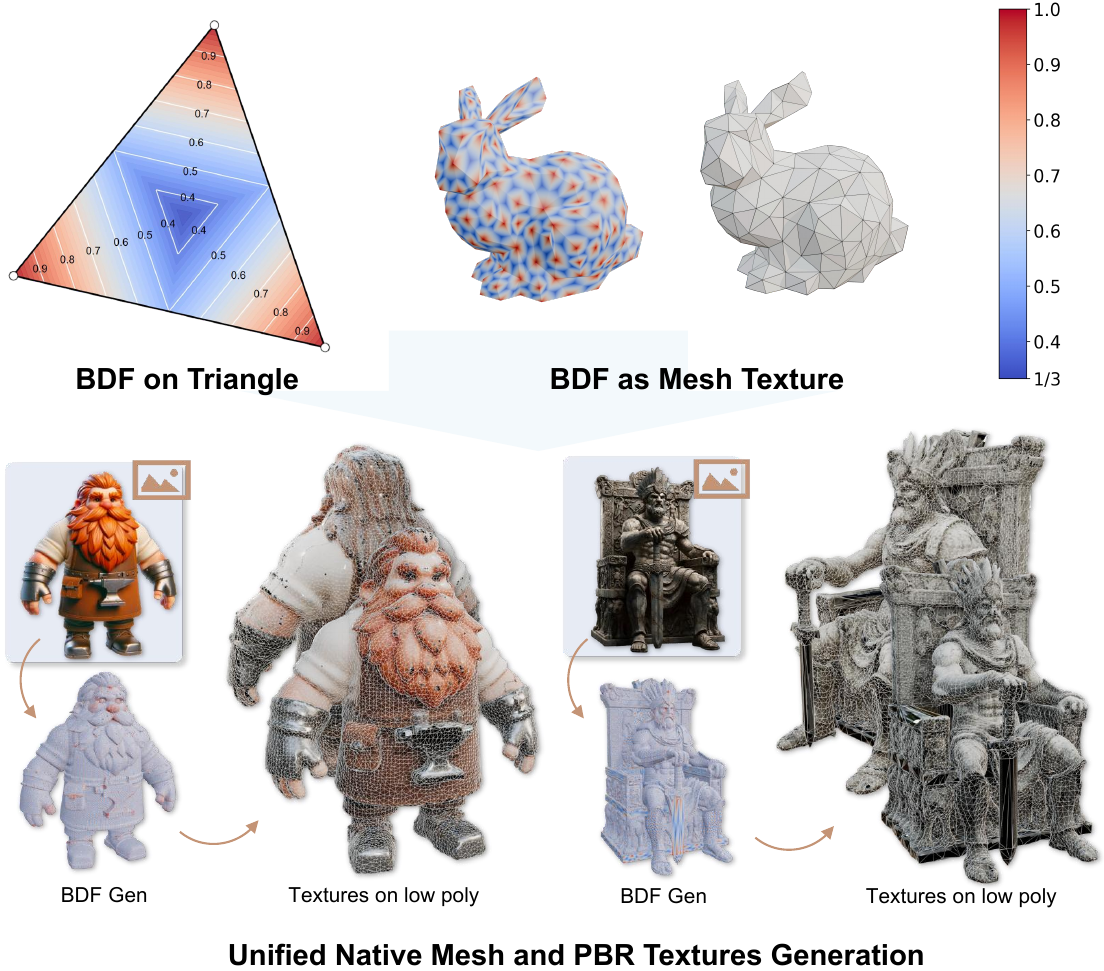}
    \caption{\textit{(Top)}: Illustration of the Barycentric Dominance Field (BDF), a scalar field with range $[1/3,1]$ defined directly on a triangular mesh surface. The white lines indicate its contour lines. \textit{(Bottom)}: Benefited from BDF representation, we achieved unified generation for Native Mesh and PBR Textures within a single framework.}
    \label{fig:bdf}
\end{figure}

\begin{abstract}
  
  Autoregressive (AR) modeling has recently achieved remarkable progress in native 3D mesh generation, largely due to its natural ability to handle variable-length, discrete data structures. However, the inherent constraints of the AR paradigm severely restrict the generated meshes, leading to limited face counts, bounded vertex resolutions, and difficulties in supporting textures. To overcome these bottlenecks, we propose the Barycentric Dominance Field (BDF), a continuous representation defined on triangular mesh surfaces that elegantly encodes vertex topological connectivity. BDF bridges the fundamental gap between discrete mesh topology and continuous diffusion-based generative modeling by transforming connectivity into a continuous surface signal. As an intrinsic mesh property, BDF shares strong similarities with texture maps, enabling its seamless integration into existing 3D diffusion pipelines without requiring architectural modifications. Extensive experiments demonstrate that BDF empowers diffusion models to generate native meshes with significantly higher quality, greater scalability, and stronger robustness compared to state-of-the-art autoregressive methods.
\end{abstract}

\section{Introduction}

Recent years have witnessed impressive progress in native 3D mesh generation. Unlike traditional pipelines that extract meshes from implicit shapes via Marching Cubes \citep{mc}, native generation directly predicts vertex coordinates and topological connections. This paradigm yields lightweight meshes with high-quality topology, significantly facilitating downstream applications such as gaming and cinematic production.

Currently, the dominant paradigm for native mesh generation relies on autoregressive (AR) models \citep{nash2020polygen,siddiqui2024meshgpt} rather than diffusion models \citep{jun2023shap, nichol2022point}. This discrepancy stems from the underlying data structures: AR models naturally handle variable-length, discrete data (such as mesh graphs), whereas diffusion models typically require fixed-length, continuous representations. Despite their tremendous success, AR-based methods suffer from inherent limitations dictated by the paradigm itself:
\textbf{(1) Restricted scalability:} The inherent upper bound on token sequence length strictly limits the maximum number of generated faces, while vocabulary size constraints inherently restrict the quantization resolution of vertices within the normalized coordinate space.
\textbf{(2) Lack of attribute support:} Incorporating other 3D attributes, such as color, UV mapping, or physically based rendering (PBR) materials, into the AR generation process remains an open challenge.
\textbf{(3) Fragile robustness:} Unlike natural language, meshes are highly sensitive to erroneous tokens. A single mispredicted token can cause severe geometric artifacts, such as surface holes or flying faces, a vulnerability that is particularly exacerbated during long-sequence generation.
Furthermore, AR models face practical bottlenecks, including slow inference speeds and high training costs. These challenges compel us to rethink the feasibility of adapting the highly successful diffusion paradigm for native mesh generation.

The crux of resolving this lies in establishing an equivalent continuous field representation for discrete triangular meshes. Such a representation would allow us to leverage the strong capability of neural networks in fitting continuous signals \citep{hornik1989multilayer}, thereby seamlessly integrating native mesh generation into mature and powerful 3D diffusion frameworks.

To this end, we propose Barycentric Dominance Field (BDF), a novel scalar field defined on the surface of triangular meshes. Formally, it is constructed simply by extracting the maximum barycentric coordinate of points on the triangle surface. As an intrinsic property of the mesh, BDF exhibits excellent mathematical properties, including $C^0$ continuity and Lipschitz continuity. These properties establish a solid foundation for successful neural network fitting.

From a 3D generation perspective, BDF is highly analogous to a texture map. As shown in Fig. \ref{fig:bdf}, this crucial insight allows us to process BDF using similar texture techniques, seamlessly incorporating it into existing VAE and diffusion frameworks. Conceptualizing BDF as a texture map elegantly decouples the geometry (shape) and topology of the mesh. The former can be obtained by leveraging powerful priors from existing 3D shape generation  \citep{xiang2025native}, allowing the generative network to focus exclusively on the latter—namely, the precise positioning of vertices on the high-resolution surface and their local connectivity.

In summary, our main contributions are as follows:

\begin{itemize}
    \item We propose BDF, a novel scalar field defined on the triangular mesh surface, and formally prove its mathematical properties. This representation bridges the gap between discrete native mesh generation and the continuous diffusion paradigm.
    \item By conceptualizing BDF as a texture map, we seamlessly integrate BDF into the powerful 3D VAE and Diffusion generative paradigm without requiring modifications to the underlying model architecture.
    \item Extensive experiments demonstrate the effectiveness of this novel mesh representation, showcasing superior generation quality, scalability, and robustness compared to state-of-the-art autoregressive methods.
\end{itemize}

\section{Related Works}

\subsection{Non-Native Mesh Generation}
To fit the continuous nature of diffusion models, most 3D generation methods encode meshes into implicit fields like Signed Distance Fields (SDFs) \citep{v1,v3,v4,v5,v7,v8,v10}. While preserving overall geometry, this process entirely discards original vertex positions and topological connectivity. Extracting explicit surfaces from these fields requires algorithms like Marching Cubes (MC). Although recent high-fidelity methods predict shapes directly on MC or Dual Contouring (DC) grids \citep{li2025sparc3d, he2025sparseflex, ju2002dual, xiang2025native, luo2025faithful}, they inevitably yield excessively dense and irregular triangulations (i.e., messy wireframes).

\subsection{Native Mesh Generation}
Inspired by NLP, autoregressive (AR) paradigms have been widely adapted for native 3D mesh generation \citep{nash2020polygen, siddiqui2024meshgpt}. To scale face counts or enhance topology, existing works explore novel tokenization strategies \citep{chen2024meshanythingv2, tang2024edgerunner, weng2024scaling, songtopology}, architectural modifications \citep{hao2024meshtron}, or post-training techniques \citep{zhao2025deepmesh,liu2025quadgpt}. However, diffusion-based native mesh generation remains sparse due to the discrete nature of meshes. Pioneering works like PolyDiff \citep{alliegro2023polydiff} and PartDiffuser \citep{yang2025partdiffuser} adapt continuous diffusion into discrete modes, while MeshCraft \citep{he2025meshcraft} designs a specialized VAE tailored for flow models. Notably, concurrent work LATO \citep{zhao2026lato} proposes an engineered mesh-to-voxel codec to enable the use of mature 3D VAEs and flow models.

Unlike LATO's heuristically designed codec, our proposed BDF leverages an intrinsic mathematical property of triangular surfaces. It establishes a natural equivalence between discrete meshes and continuous scalar fields, seamlessly integrating with the Dual Contouring algorithm. This provides a mathematically grounded, elegant, and highly generalizable solution for native mesh generation.

\section{Barycentric Dominance Field}
\label{sec:bdf}

We introduce the Barycentric Dominance Field (BDF), a novel scalar field defined directly on the triangle mesh surface. 

\subsection{Formulation}
For a given triangle $T$ with vertices $V_1, V_2, V_3$, any point $p \in T$ can be uniquely expressed using its barycentric coordinates $(w_1, w_2, w_3)$ as follows:
\begin{gather*}
    p =  w_1 V_1 + w_2 V_2 + w_3 V_3 \\
    \text{where}\quad w_1+w_2+w_3 = 1,
    \quad w_i \ge 0 \quad \text{for} \quad i \in \{1,2,3\}
\end{gather*}
We define the Barycentric Dominance Field (BDF) value at point $p$, denoted as $B(p)$, as the maximum of its barycentric coordinates:
\begin{equation*}
    B(p) = \max(w_1, w_2, w_3)
\end{equation*}
This definition naturally extends to a complete 3D mesh surface $\mathcal{M} = \{T_k\}$. For any point on the mesh, its BDF value is evaluated based on the specific triangle it resides in, as shown in Fig. \ref{fig:bdf}.
\subsection{Theoretical Properties}
\label{sec:bdf_properties}

The proposed BDF has several simple yet useful properties for neural surface representation:

\textbf{$C^0$ continuity.}
The BDF is continuous over the mesh surface. Intuitively, within each triangle, the barycentric coordinates change smoothly as the point moves on the triangle, and thus their maximum also changes continuously. Moreover, when a point moves across a shared edge between two adjacent triangles, the barycentric coordinate corresponding to the opposite vertex becomes zero on both sides. Therefore, the BDF value computed from either triangle is consistent on the shared edge, which avoids abrupt jumps across triangle boundaries.

\textbf{Lipschitz continuity.}
The BDF is also Lipschitz continuous on a finite non-degenerate triangular mesh. In simple terms, this means that the BDF value cannot change arbitrarily fast when the point moves along the surface. This property comes from the fact that barycentric coordinates are affine functions within each triangle, and the maximum operation does not amplify the variation of its inputs. As a result, the BDF has bounded local variation over the mesh surface.

These properties make BDF a well-behaved supervision signal for neural field learning. Its continuity avoids artificial discontinuities on the mesh, while its Lipschitz continuity provides a bounded and stable target for function approximation. Therefore, BDF can be effectively fitted by standard neural networks. We provide detailed proofs of these properties in Appendix~\ref{app:bdf_proof}.

\section{Method}

\label{sec:method}

To realize native mesh generation based on BDF, we adopt Trellis.2 \citep{xiang2025native}—a powerful 3D generation pipeline based on sparse voxels and flow models—as our backbone. We first introduce the algorithm design for mesh-to-voxel encoding and decoding, followed by the generation pipeline.

\subsection{Algorithms of BDF Encoding and Decoding}

\begin{figure}[tbp]
    \centering
    \includegraphics[width=1\textwidth]{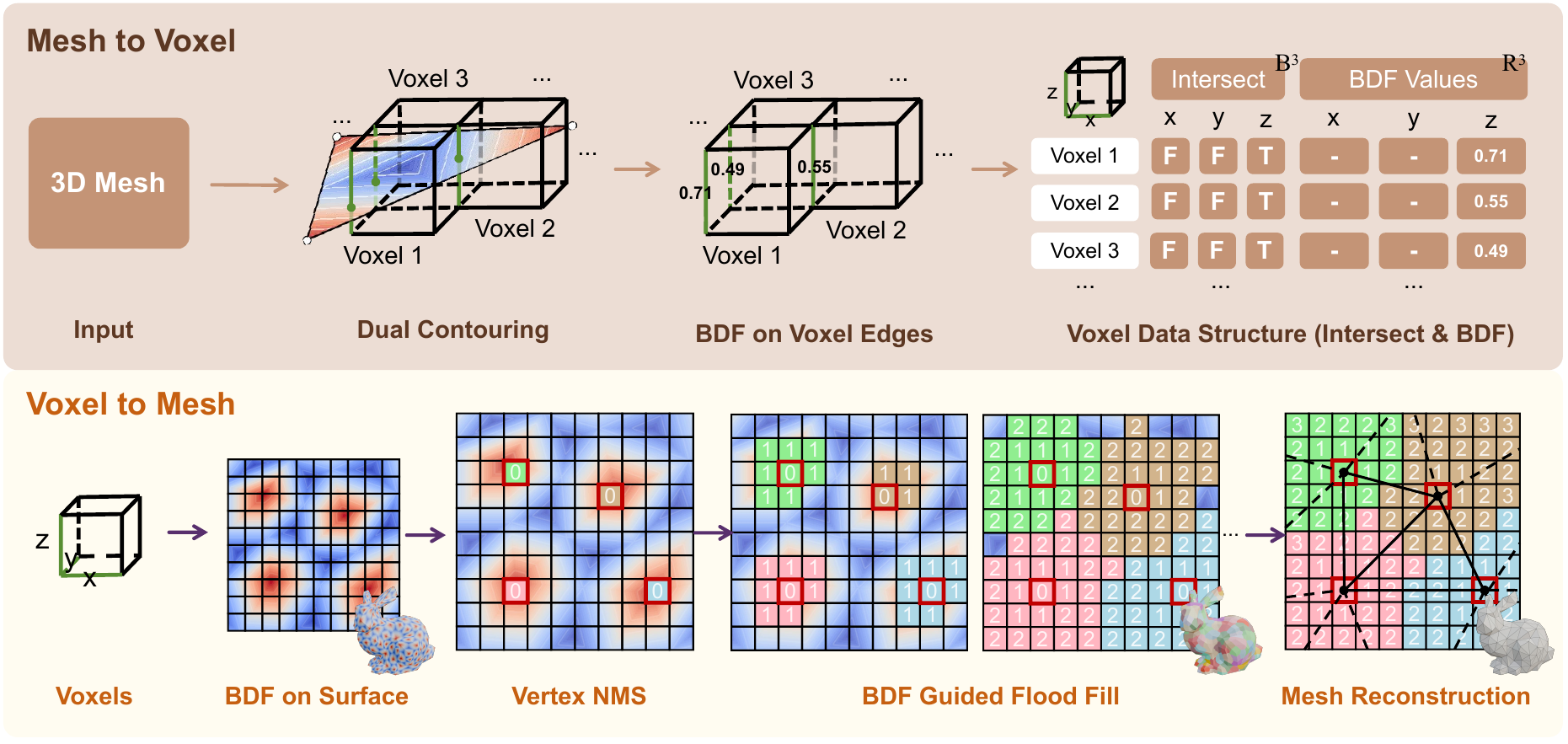}
    \caption{\textbf{BDF Encoding and Decoding Pipeline.} \textit{Top (Encoding):} Voxel edges intersecting the mesh surface are highlighted in green, directly storing the BDF values of the intersection points. Since one edge is shared among four surrounded voxels, to prevent data redundancy, each sparse voxel stores only three orthogonal edges (marked as $x, y, z$). \textit{Bottom (Decoding):} A 2D toy example of our Flood-Fill algorithm. Mesh vertices are first identified via Non-Maximum Suppression (NMS) on local BDF maxima. The flood-fill process then propagates along BDF contours to construct an equivalent Voronoi-style mesh representation.}
    \label{fig:pipe}
\end{figure}

\textbf{Dual Contouring.} To efficiently compress complex mesh geometries into sparse voxels, Trellis.2 utilizes a training-free, explicit mesh-to-sparse-voxel compression scheme based on the Dual Contouring (DC) algorithm. Specifically, given an input mesh and a high-resolution voxel grid, the DC algorithm first determines whether the voxel edges along the $x$, $y$, and $z$ axes intersect with the mesh triangles, yielding three bool intersection flags $\in \text{B}^3 $. For each intersected edge, it assigns a small quad whose vertices are the centers of the four surrounding voxels (termed Dual Vertices $\in \text{R}^3$). A Quadratic Error Function (QEF) is then minimized to fit these four vertices to the underlying surface, with a regularization term restricting them within their respective voxels. To compute the QEF, Hermite data (i.e., intersection points and surface normals) are temporarily cached on each intersected edge. Additionally, following the design of FlexiCubes \citep{shen2023flexible}, each dual vertex stores an extra splitting scalar ($\in \text{R}^1$) to indicate the subdivision pattern of the quad.

\textbf{Voxel Encoding.} We retain the original DC algorithm entirely and introduce an incremental extension: during the computation of Hermite data, we additionally calculate the barycentric coordinates of the intersection points and store the corresponding BDF values on the voxel edges. As illustrated in Fig. \ref{fig:pipe}, given the sufficiently high resolution of the voxel grid (typically $512^3$ or $1024^3$), this encoding process can be viewed as a dense sampling of the surface BDF. To explicitly encourage the network to learn vertex positions, we further match the original mesh vertices with the intersection points in the Hermite data, assigning a peak BDF value of $1.0$ to the corresponding edges.

\textbf{Surface Decoding.} Given the voxel data where each edge encodes a BDF value, we first decode the surface quads directly using dual vertices and intersection flags. Since each quad corresponds to an intersected edge, it naturally inherits a BDF value. To facilitate subsequent algorithmic processing, we transform these BDF-augmented quads into a dual graph structure: each quad center acts as a graph node endowed with its BDF value, and edges are formed by connecting adjacent quads.

\textbf{Mesh Reconstruction.} We first apply Non-Maximum Suppression (NMS) on the dual graph to identify the mesh vertices. Inspired by the concept of dual meshes, we design a BDF-guided flood-fill algorithm to construct an equivalent Voronoi-style representation for mesh reconstruction. The "flood" initiates simultaneously from all identified mesh vertices and propagates along the BDF contours. When floodfronts from different vertices meet, an edge is instantiated between the corresponding vertices. Upon completion, this process yields a wireframe structure composed of vertices and edges. Since this wireframe inherently adheres to the mesh surface, we simply traverse it to find simple cycles of length 3 or 4 to reconstruct the final, complete mesh faces. As shown in Fig. \ref{fig:pipe}, we provide a 2D toy example of this flood-fill algorithm, where each pixel denotes a DC quad. Intuitively, during the flood-fill process, fronts meet at BDF saddle points (where BDF value is $1/2$) to recover the original mesh edges, and the process terminates at the local minima (where BDF value is $1/3$). The pseudocode for this algorithm is provided in the Appendix \ref{sec:appendix_flood_fill}.

\begin{figure}[tbp]
    \centering
    \includegraphics[width=1\textwidth]{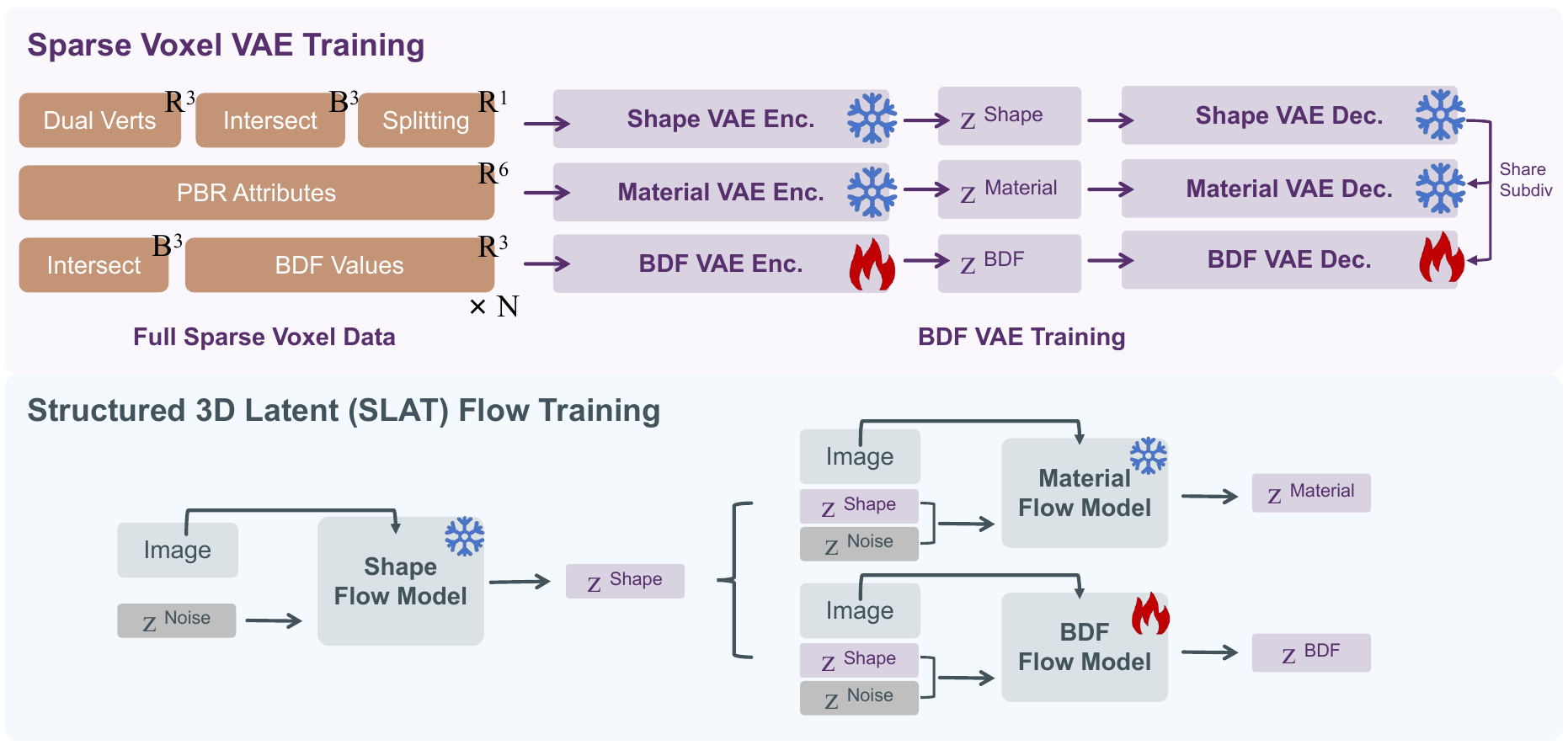}
    \caption{\textbf{Architecture of the Decoupled VAE and Flow Models.} The Shape and Material VAEs share an identical U-Net-style architecture, differing only in input/output channels (7 and 6, respectively). We directly repurpose this architecture for the BDF VAE with 3 channels. A similar decoupling strategy is applied to the Flow Models.}
    \label{fig:pipegen}
\end{figure}

\subsection{Generation Pipeline}

\textbf{VAE.} As shown in Fig. \ref{fig:pipegen}, since the BDF encoding aligns closely with how Trellis.2 encodes dual vertices and textures, it only incurs a marginal cost of 3 additional float values per sparse voxel, representing the BDF values on the $x$, $y$, and $z$ edges. Considering that the Shape and Material (Texture) modules in Trellis.2 utilize identical U-Net-style \citep{u-vae} VAEs with input/output dimensions of 7 and 6 respectively, we directly repurpose this architecture to construct the BDF VAE, setting the input/output dimensions to 3. Notably, the intersection flags can be directly retrieved from the Shape VAE, eliminating the need for redundant encoding of BDF VAE. 

Our VAE design ensures complete decoupling from the original Shape and Material VAEs, sharing only essential structural information (e.g., intersections and subdivision patterns during voxel upsampling). This design elegantly leverages the powerful shape priors from the Shape VAE to construct the underlying surface, allowing the BDF VAE to focus exclusively on learning the precise locations of mesh vertices on this high-fidelity surface and their local connectivity. 

During training, to eliminate background noise and focus the network's capacity on the actual surface, the reconstruction loss is exclusively computed on the valid intersected edges, denoted as the set $E_{int}$. Furthermore, to encourage the network to predict sharp and accurate vertices, we introduce a spatially varying weight $w_i$. The training loss for the BDF VAE is formulated as:

$$ \mathcal{L}_{VAE} = \lambda_{field} \mathcal{L}_{field} + \lambda_{KL} \mathcal{D}_{KL} $$

where $\mathcal{D}_{KL}$ regularizes the latent space, and the field reconstruction loss $\mathcal{L}_{field}$ is defined as a weighted Binary Cross-Entropy (BCE) loss over the intersected edges:

$$ \mathcal{L}_{field} = - \frac{1}{\sum_{i \in E_{int}} w_i} \sum_{i \in E_{int}} w_i \left[ x_i \log(\hat{x}_i) + (1 - x_i) \log(1 - \hat{x}_i) \right] $$

Here, $x_i$ and $\hat{x}_i$ represent the normalized ground truth and predicted BDF values, respectively. The weight $w_i$ is defined as:

$$ w_i = \begin{cases} \lambda_{point}, & \text{if } x_i > 0.999 \\ 1, & \text{otherwise} \end{cases} $$

By assigning a significantly larger weight ($\lambda_{point} \gg 1$) to the actual mesh vertices ($x_i > 0.999$), we effectively penalize vertex blurring.

\textbf{Flow Model.} Following the training and inference paradigm of the Material Flow Model in Trellis.2, we first employ the Shape Flow Model to predict the shape latents $z^{\text{shape}}$ conditioned on the input image. Subsequently, a sparse Diffusion Transformer (DiT) \citep{dit} predicts the BDF latents $z^{\text{BDF}}$, conditioned jointly on the input image and the generated shape latents $z^{\text{shape}}$. As illustrated in Fig. \ref{fig:pipegen}, similar to the VAE stage, we completely freeze the Shape and Material Flow Models, decoupling them from the BDF Flow. This strategy fully preserves the robust shape and texture generation capabilities of the pre-trained models.

\begin{figure}[tbp]
    \centering
    \includegraphics[width=1\textwidth]{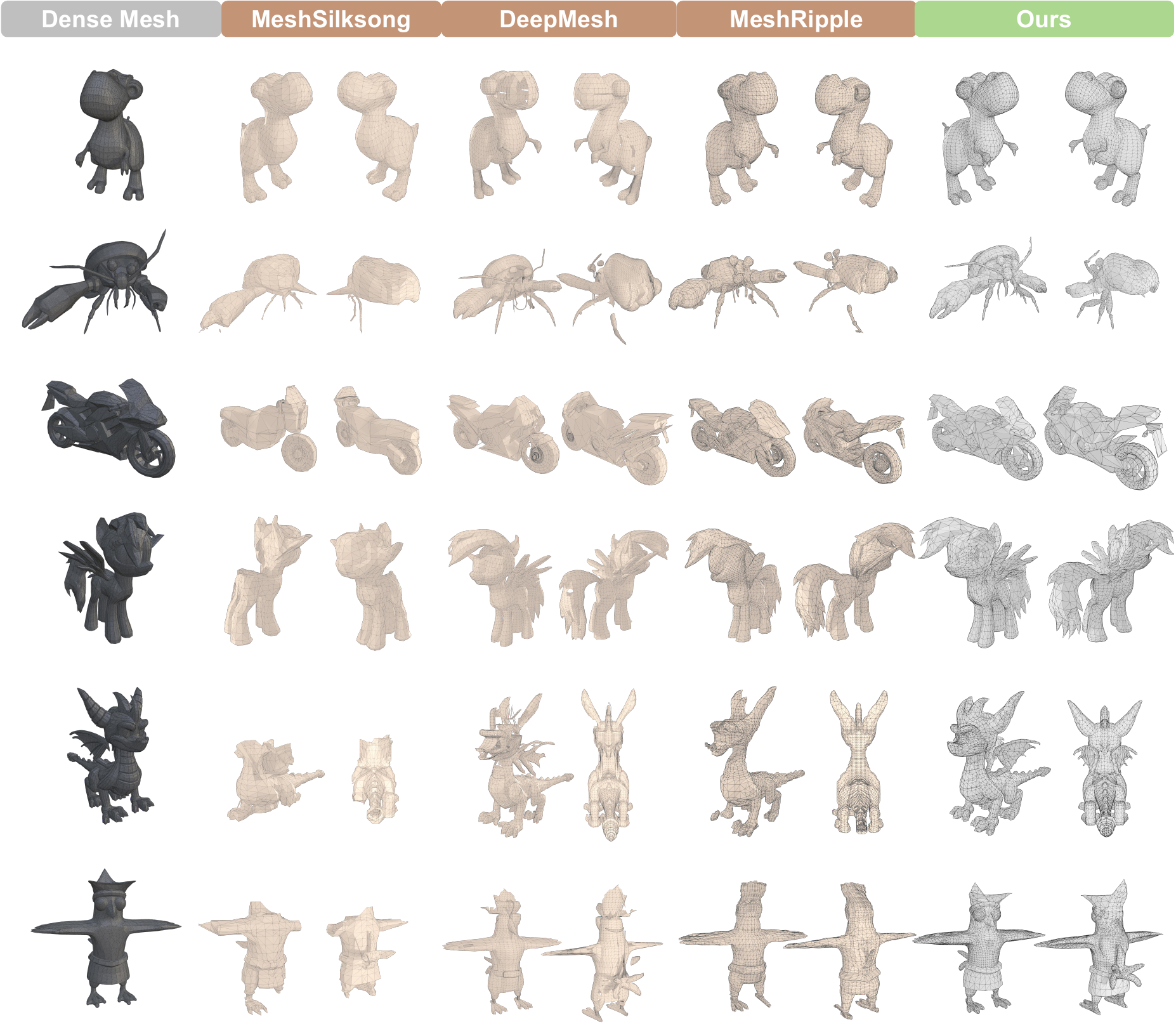}
    \caption{\textbf{Qualitative Comparison on the Toys4k Dataset.} Autoregressive (AR) baselines struggle with complex geometries. MeshSilksong frequently hits its generation upper bound due to the strict 10k maximum token limit. Although DeepMesh and MeshRipple extend this limit to 90k, the risk of predicting erroneous tokens increases significantly with sequence length, inevitably leading to fragmented and broken faces. In contrast, our method elegantly bypasses the inherent face count and robustness limitations of AR paradigms, consistently delivering stable and high-quality topologies.}
    \label{fig:exps}
\end{figure}

\begin{figure}[tbp]
    \centering
    \includegraphics[width=1\textwidth]{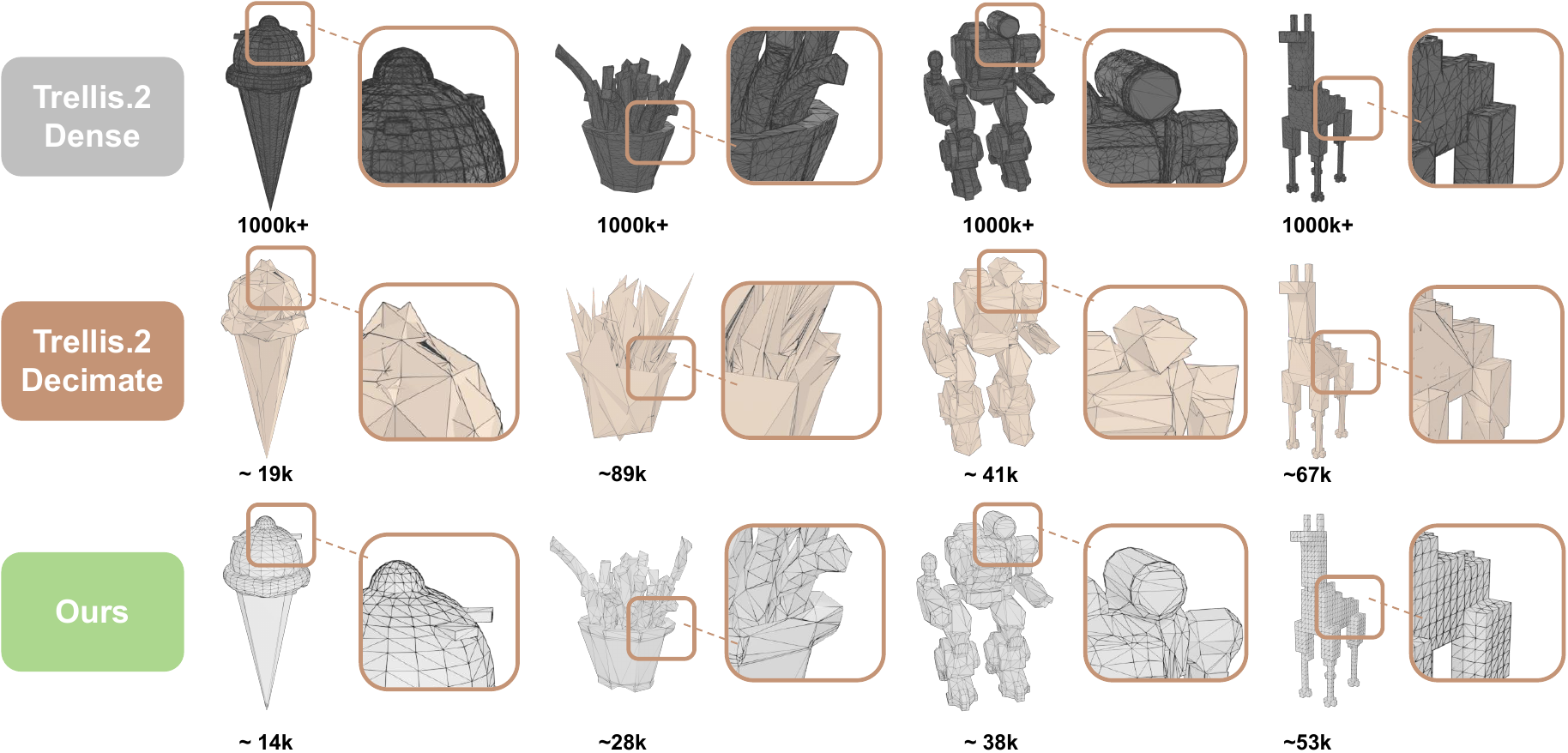}
    \caption{\textbf{Comparison with Decimated Dense Meshes.} We directly compare our method with Trellis.2, a representative state-of-the-art dense mesh generator. When subjected to standard mesh decimation to meet downstream requirements, the ultra-high-poly outputs of Trellis.2 (typically $>1000\text{k}$ faces) suffer from severe geometric degradation and chaotic topology. In contrast, our approach directly yields meshes with moderate face counts and improved connectivity patterns.}
    \label{fig:comptrellis}
    \vspace{-10pt}
\end{figure}

\section{Experiments}
\begin{table}[t]
\caption{\textbf{Quantitative Results on the Toys4k Dataset.} Our method consistently outperforms representative autoregressive baselines across all geometric metrics. Notably, as the object complexity scales from low-poly ($0 \sim 10\text{k}$) to high-poly ($0 \sim 40\text{k}$), our approach maintains robust performance, whereas AR methods suffer significant degradation.}
\label{tab:main}
\centering
\begin{tabular}{lccccccc}
\toprule
\multirow{2}{*}{Methods} & \multicolumn{3}{c}{Toys4k (0 $\sim$ 10 k)}                      & \multicolumn{3}{c}{Toys4k (0 $\sim$ 40 k)}                      \\ \cmidrule(lr){2-4}\cmidrule(lr){5-7}
                         & CD ($\times e^{-3}$) $\downarrow$ & HD $\downarrow$             & |NC| $\uparrow$          & CD ($\times e^{-3}$) $\downarrow$  & HD $\downarrow$             & |NC| $\uparrow$          \\ \midrule
MeshSilksong            & 14.85                      & 0.254          & 0.872          & 41.247                     & 0.411          & 0.790           \\
DeepMesh                 & 17.84                      & 0.243          & 0.889          & 25.113                     & 0.331          & 0.826          \\
MeshRipple               & 3.830                       & 0.154          & 0.895          & 15.584                     & 0.299          & 0.818          \\
Ours                     & \textbf{0.102}             & \textbf{0.038} & \textbf{0.929} & \textbf{0.123}             & \textbf{0.051} & \textbf{0.884} \\ \bottomrule
\end{tabular}
\end{table}

\subsection{Datasets and Implementation Details}
Due to storage constraints, we utilize the Texverse \citep{zhang2025texverse} dataset and filter a high-quality subset of approximately 180k 3D assets to train the VAE and flow models. The BDF VAE is trained on 8 $\times$ H200 GPUs with a batch size of 8, taking roughly 16 hours. Similarly, the BDF Flow model is trained on 8 $\times$ H200 GPUs with a batch size of 8, requiring about 24 hours. Following the training protocol of Trellis.2, all models are optimized using AdamW \citep{adaw} (learning rate $1 \times 10^{-4}$, weight decay $0.01$) and employ classifier-free guidance with a drop rate of $0.1$.

\subsection{Evaluation Setting}
\textbf{Datasets and Metrics.} We select the Toys4k \citep{toys4k} dataset to compare our method against state-of-the-art baselines. To rigorously evaluate the robustness of our approach on high-poly, artist-created meshes (typically $>10\text{k}$ faces), we construct two evaluation splits: a standard set of 200 meshes with $0 \sim 10\text{k}$ faces, and a complex set of 100 meshes with $0 \sim 40\text{k}$ faces. We employ three standard geometric metrics: Chamfer Distance (CD), Hausdorff Distance (HD), and Normal Consistency ($|\text{NC}|$), calculated using $30,000$ sampled surface points.

\textbf{Baselines and Condition Alignment.} We select three representative autoregressive (AR) native mesh generation models as our baselines: MeshSilksong \citep{songtopology}, DeepMesh \citep{zhao2025deepmesh}, and MeshRipple \citep{lin2025meshripple}. Since these AR models require point clouds as input conditions, we design a fair alignment pipeline: we first feed the rendered images of artist-created meshes into Trellis.2 to generate dense meshes, and then sample point clouds from these surfaces to serve as the condition for the AR baselines. Then we compare the metrics of generated mesh with Trellis.2 dense mesh. It is notable that during the generation of Trellis.2 dense mesh, the shape latent of trellis.2 is extracted as our BDF flow model condition to generate BDF latent.

\begin{figure}[t]
    \centering
    \includegraphics[width=1\textwidth]{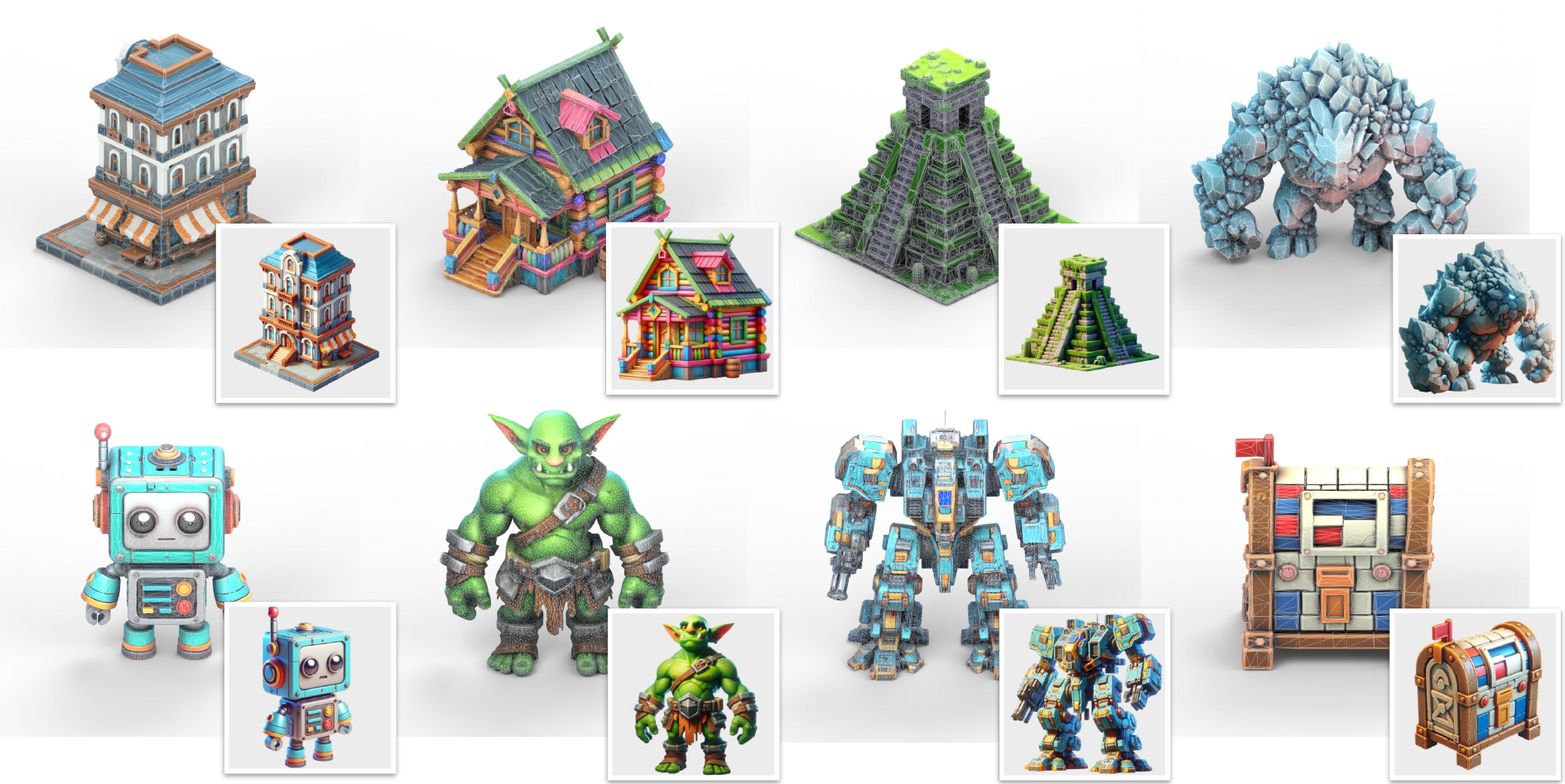}
    \caption{\textbf{Unified generation of mesh topology and PBR textures.} The mesh topology is indicated with white lines, please zoom in on the rendering results to view the topology details.}
    \label{fig:pbr}
\end{figure}

\begin{table}[t]
\caption{\textbf{Quantitative Results compared to three different QEM decimation methods on Trellis.2 dense mesh.} We sampled our 100 Toys4k generation results conditioned on image whose face counts are around 8K–12K, and decimate the corresponding Trellis.2 dense mesh during generation to the same face count as ours with three QEM methods, and then compute CD, HD, and |NC| against the original dense mesh. The "Ref GT" denotes the dense mesh without decimation, sampled into a point cloud and compared against itself; it can be regarded as an upper-bound reference for the metrics.}
\label{tab:qem3}
\centering
\begin{tabular}{ccccccc}
\toprule
QEM Methods & CuMesh & Open3D & Blender & Ours(512) & Ours(1024 cascade) & Ref GT \\ \midrule
CD ($\times e^{-3}$) $\downarrow$    & 0.201             & 0.112  & 0.144   & 0.082     & \textbf{0.075}     & 0.052  \\
HD $\downarrow$           & 0.077             & 0.053  & 0.067   & 0.035     & \textbf{0.031}     & 0.015  \\
|NC| $\uparrow$        & 0.932             & 0.940  & 0.920   & 0.942     & \textbf{0.949}     & 0.973  \\ \bottomrule
\end{tabular}
\end{table}

\begin{table}[t]
\caption{\textbf{Ablation Study on VAE Loss and Architecture Design.} The vertex reconstruction accuracy is evaluated with F-scores under varying matching tolerances, where $V$ denotes the voxel length. The results highlight the superiority of the BCE loss and the necessity of the decoupled architecture.}
\label{tab:abl}
\centering
\begin{tabular}{lccccccc}
\toprule
\multirow{2}{*}{Ablation} & \multirow{2}{*}{CD ($\times e^{-3}$) $\downarrow$} & \multirow{2}{*}{HD $\downarrow$} & \multicolumn{4}{c}{F-score $\uparrow$}                                       \\ \cmidrule(lr){4-7} 
                          &                           &                     & 0.5V           & 1V             & 2V             & 3V             \\ \midrule
Decoupled VAE  + BCE          & \textbf{0.006}            & \textbf{0.011}      & \textbf{0.599} & \textbf{0.702} & \textbf{0.958} & \textbf{0.991} \\
Decoupled VAE + MSE            & 0.008                     & 0.013               & 0.493          & 0.643          & 0.957          & 0.989          \\
Jointed VAE + BCE               & 0.032                     & 0.031               & 0.368          & 0.516          & 0.874          & 0.942          \\ \bottomrule
\end{tabular}
\end{table}

\begin{figure}[t]
    \centering
    \includegraphics[width=0.9\textwidth]{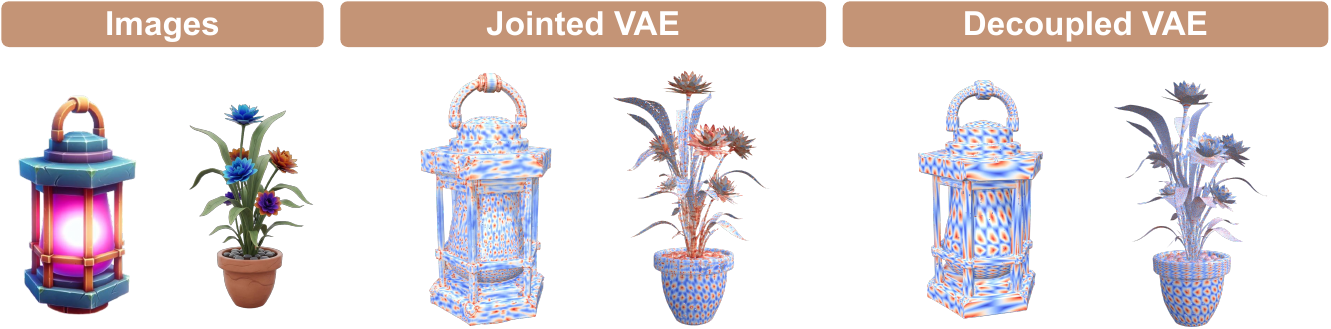}
    \caption{\textbf{Visual Ablation of Jointed vs. Decoupled VAE Designs.} We visualize the generated BDF fields decoded from different latent spaces. The Decoupled VAE produces a crisp, regular BDF. In contrast, the Jointed VAE yields a muddy field with ambiguous and chaotic peak distributions.}
    \label{fig:abl}
\end{figure}

\subsection{Results and Comparisons}

\textbf{Quantitative Results.} As shown in Table \ref{tab:main}, while AR baselines perform well on low-poly meshes ($0 \sim 10\text{k}$), their geometric metrics degrade significantly on more complex samples ($0 \sim 40\text{k}$). In contrast, our method outperforms them across all metrics and exhibits exceptional robustness as data complexity scales, particularly in CD scores. More results about inference-efficiency and other mesh distribution are provided in Appendix \ref{appendix:infer}.

\textbf{Qualitative Results.} Fig. \ref{fig:exps} illustrates that AR methods struggle with complex shapes, often failing due to token limits or producing fragmented faces. This stems from error accumulation in long sequences, to which discrete meshes are notoriously sensitive. Our method elegantly bypasses these token limits, with its resolution bounded only by the Dual Contouring grid.

\textbf{Comparison to decimation of dense mesh.}
Furthermore, we compare our approach against ultra-high-poly generation (Trellis.2) followed by naive decimation with Blender in Fig. \ref{fig:comptrellis} and with three different QEM algorithms in Tab. \ref{tab:qem3}. While Trellis.2 yields intricate meshes, their massive face counts ($>1$M) necessitate decimation for practical downstream deployment, which inevitably causes severe geometric degradation and chaotic topology. Conversely, our method directly outputs meshes with moderate face counts and improved connectivity patterns. After decimation, the metrics of trellis.2 degrade relative to the original dense mesh, and all are worse than our results. The quantitative and qualitative results jointly demonstrate the importance of our method: for a dense mesh, it is difficult for existing QEM decimation algorithms to balance post-decimation precision and wiring quality simultaneously. Therefore, directly learning the native vertices and topological connections from the dataset as our method is much suitable for downstream 3D applications.

\textbf{Unified generation of topology and PBR.} Finally, Fig. \ref{fig:bdf} and \ref{fig:pbr} demonstrates the unified generation of mesh topology and PBR textures. To our knowledge, we are the \textbf{first} to achieve this unified generation within a single framework, highlighting the remarkable scalability of BDF.

\subsection{Ablation Studies}

\textbf{VAE Loss: BCE vs. MSE.} While BDF resembles a texture map (suggesting MSE loss), its sharp nature (peaking at $1.0$ at vertices) requires precise localization. BCE loss provides stronger gradient signals to restore these sharp peaks. As shown in Table \ref{tab:abl}, we evaluate vertex reconstruction using CD, HD, and F-score (under strict tolerances of $0.5, 1, 2, 3$ voxel units). Although MSE yields comparable CD and HD, it falls significantly short under the stricter F-score, proving BCE is crucial for accurate vertex localization.

\textbf{Decoupled vs. Jointed VAE.} Since BDF correlates with surface geometry (e.g., peaks at sharp features), integrating it into the Shape VAE (expanding channels from 7 to 10) and fine-tuning via ControlNet-style zero-convolutions \citep{zhang2023adding} might seemingly facilitate learning. However, Table \ref{tab:abl} shows this Jointed VAE unexpectedly entangles the latent space, severely degrading vertex accuracy. Visually (Fig. \ref{fig:abl}), flow models trained on this joint latent space generate "muddy" BDF fields with chaotic peak distributions, confirming the necessity of our decoupled architecture.

\section{Conclusion}

In this paper, we introduced the Barycentric Dominance Field (BDF), a continuous scalar field that elegantly bridges discrete native mesh generation and the continuous diffusion paradigm. By conceptualizing BDF as an intrinsic surface texture, we seamlessly integrated it into existing 3D VAE and diffusion frameworks without architectural modifications. This design bypasses the token-length and robustness bottlenecks of autoregressive models, enabling the first unified generation of native meshes and PBR textures. Experiments demonstrate our method's state-of-the-art geometric quality and scalability, particularly on complex, high-poly assets.

\textbf{Limitations and Future Work.} While effective, BDF's current codec need not be strictly confined to Dual Contouring (DC). During encoding, the DC grid resolution imposes a physical limit: topological information of extremely small faces missing grid intersections is inevitably lost. During decoding, because our flood-fill algorithm operates on the DC-recovered surface, self-intersecting triangles in the original mesh can mislead propagation, yielding erroneous connections. Future work will explore resolution-independent codecs to fully unleash BDF's potential for intricate, ultra-fine 3D structures.

\section{Acknowledgements}
This work is partially supported by funding and computing resources provided by the HKU-Shanghai
Intelligent Computing Research Center (H200) and partially supported by the Shenzhen Loop Area Institute. Also, this work was supported by the General Research Fund of the Research Grants Council of Hong Kong under Grant No. 17200725 and the JC STEM Lab funded
by The Hong Kong Jockey Club Charities Trust. The authors declare no competing interests.

\bibliographystyle{plainnat}
\bibliography{ref}







\newpage
\appendix

\section*{Technical appendices and supplementary material}

\section{Proof of BDF Continuity and Lipschitz Continuity}
\label{app:bdf_proof}

In this section, we provide the detailed proof for the continuity properties of the proposed Barycentric Dominance Field (BDF).

\subsection{$C^0$ Continuity}

\noindent\textbf{Proposition 1.}
Let $\mathcal{M}$ be a triangular mesh surface, and let $B(p)$ be the BDF value defined as the maximum barycentric coordinate of point $p$. Then $B(p)$ is $C^0$ continuous on $\mathcal{M}$.

\noindent\textbf{Proof.}
We first consider a single triangle $T$ with vertices $V_1,V_2,V_3$. For any point $p \in T$, its barycentric coordinates $(w_1,w_2,w_3)$ satisfy
\begin{equation}
    p = w_1 V_1 + w_2 V_2 + w_3 V_3,
    \quad
    w_1+w_2+w_3 = 1.
\end{equation}
Inside a non-degenerate triangle, each barycentric coordinate $w_i$ is an affine function of $p$. Therefore, $w_1,w_2,w_3$ are continuous functions on $T$. Since the maximum of finitely many continuous functions is still continuous, we have
\begin{equation}
    B(p) = \max(w_1,w_2,w_3)
\end{equation}
continuous inside each triangle.

It remains to verify continuity across triangle boundaries. Consider two adjacent triangles $T_a$ and $T_b$ sharing an edge with endpoints $V_1$ and $V_2$. Let their third vertices be $V_3$ and $V_4$, respectively. For any point $p$ on the shared edge, $p$ can be represented as
\begin{equation}
    p = w_1 V_1 + w_2 V_2,
    \quad
    w_1 + w_2 = 1,
    \quad
    w_1,w_2 \ge 0.
\end{equation}
When $p$ is viewed as a point in triangle $T_a$, its barycentric coordinate associated with $V_3$ is zero. Thus, its barycentric coordinates in $T_a$ are $(w_1,w_2,0)$. Similarly, when $p$ is viewed as a point in triangle $T_b$, its barycentric coordinate associated with $V_4$ is zero, and the coordinates are also $(w_1,w_2,0)$ with respect to the shared edge.

Therefore, the BDF value computed from either side is
\begin{equation}
    B(p) = \max(w_1,w_2,0) = \max(w_1,w_2).
\end{equation}
Hence, the BDF values agree on shared edges. The same argument naturally holds at shared vertices, where one barycentric coordinate is one and the others are zero. Consequently, $B(p)$ is continuous across all triangle boundaries, and thus is $C^0$ continuous on the whole mesh surface $\mathcal{M}$.
\hfill$\square$

\subsection{Lipschitz Continuity}

\noindent\textbf{Proposition 2.}
Let $\mathcal{M}$ be a finite triangular mesh composed of non-degenerate triangles. Then the BDF $B(p)$ is Lipschitz continuous on $\mathcal{M}$ with respect to the intrinsic surface distance.

\noindent\textbf{Proof.}
We first prove the Lipschitz continuity on a single triangle $T$. Let $T$ have vertices $V_1,V_2,V_3$. Any point $p \in T$ can be written as
\begin{equation}
    p = V_1 + u_1 (V_2 - V_1) + u_2 (V_3 - V_1),
\end{equation}
where $u_1 = w_2$ and $u_2 = w_3$. The remaining barycentric coordinate is
\begin{equation}
    w_1 = 1 - u_1 - u_2.
\end{equation}
Since $T$ is non-degenerate, the mapping from $p$ to $(u_1,u_2)$ is affine on the triangle plane. Therefore, there exists a finite constant $L_T > 0$ such that for any $p,q \in T$,
\begin{equation}
    \|w(p)-w(q)\|_2 \le L_T \|p-q\|_2,
\end{equation}
where $w(p)=(w_1(p),w_2(p),w_3(p))$ denotes the barycentric coordinate vector of $p$.

Next, we use the fact that the maximum function is $1$-Lipschitz with respect to the $\ell_\infty$ norm. For any two vectors $x,y \in \mathbb{R}^3$, we have
\begin{equation}
    |\max_i x_i - \max_i y_i|
    \le \max_i |x_i-y_i|
    = \|x-y\|_\infty
    \le \|x-y\|_2.
\end{equation}
Applying this inequality to the barycentric coordinate vectors of $p$ and $q$, we obtain
\begin{equation}
\begin{split}
    |B(p)-B(q)|
    &= |\max_i w_i(p) - \max_i w_i(q)| \\
    &\le \|w(p)-w(q)\|_2 \\
    &\le L_T \|p-q\|_2.
\end{split}
\end{equation}
Thus, $B(p)$ is Lipschitz continuous within each triangle.

Now consider the whole mesh $\mathcal{M}=\{T_k\}_{k=1}^{N}$. Since the mesh contains a finite number of non-degenerate triangles, each triangle has a finite Lipschitz constant $L_{T_k}$. Define
\begin{equation}
    L = \max_{k=1,\dots,N} L_{T_k}.
\end{equation}
Then $L$ is finite.

For any two points $p,q \in \mathcal{M}$, consider a surface path $\gamma$ connecting them. This path can be decomposed into finitely many segments, where each segment lies inside a single triangle. Applying the per-triangle Lipschitz bound to each segment and summing over all segments gives
\begin{equation}
    |B(p)-B(q)| \le L \, \mathrm{Length}(\gamma).
\end{equation}
Taking the infimum over all surface paths $\gamma$ connecting $p$ and $q$, we obtain
\begin{equation}
    |B(p)-B(q)| \le L \, d_{\mathcal{M}}(p,q),
\end{equation}
where $d_{\mathcal{M}}(p,q)$ denotes the intrinsic surface distance on $\mathcal{M}$. Therefore, the BDF is globally Lipschitz continuous on the mesh surface.
\hfill$\square$

\section{Details of the BDF-Guided Flood-Fill Algorithm}
\label{sec:appendix_flood_fill}

In this section, we provide the detailed implementation of the BDF-guided flood-fill algorithm used for mesh reconstruction, as introduced in Section \ref{sec:method}. 

The core objective of this algorithm is to robustly extract topological connections (i.e., mesh edges) between the identified mesh vertices (seeds) on the dual graph. To achieve high efficiency and parallelism, we formulate this flood-fill process as a multi-source shortest path problem. 

Specifically, given the dual graph $\mathcal{G} = (\mathcal{V}, \mathcal{E})$ where nodes represent dual contouring quads and edges represent their adjacency, each node $v \in \mathcal{V}$ is associated with a predicted BDF value $f(v)$. Let $\mathcal{S} \subset \mathcal{V}$ denote the set of identified mesh vertices (seeds). We define the directed traversal weight (or "flow resistance") from node $u$ to an adjacent node $v$ as:

$$ w(u, v) = \exp(\alpha \cdot (f(v) - f(u))) $$

where $\alpha$ is a scaling hyperparameter. Since BDF values peak at mesh vertices ($1.0$) and monotonically decrease towards the face centers ($1/3$), a negative difference ($f(v) - f(u) < 0$) indicates "flowing downhill" along the BDF gradient, resulting in a minimal traversal cost. Conversely, flowing uphill incurs an exponentially large penalty. This design strictly constrains the flood to propagate along the BDF contours.

To solve this efficiently, we introduce a virtual \textit{Super Source} node $v_{super}$ connected to all seed nodes in $\mathcal{S}$ with zero-weight edges. This elegantly converts the multi-source propagation into a standard single-source Dijkstra's algorithm, allowing simultaneous flooding from all seeds. After computing the shortest paths, each node is assigned an "owner" corresponding to its closest seed via predecessor tracking. Finally, the original mesh edges are recovered by identifying the meeting frontiers—that is, the edges $(u, v) \in \mathcal{E}$ where the adjacent nodes belong to different owners. The complete procedure is summarized in Algorithm 1.

\begin{algorithm}[h]
\caption{BDF-Guided Flood-Fill for Mesh Reconstruction}
\label{alg:flood_fill}

\textbf{Input:} 
Dual graph \ensuremath{\mathcal{G} = (\mathcal{V}, \mathcal{E})}, 
BDF field \ensuremath{f: \mathcal{V} \rightarrow \mathbb{R}}, 
Seed nodes \ensuremath{\mathcal{S} \subset \mathcal{V}}, 
Scaling factor \ensuremath{\alpha} \\

\textbf{Output:} 
Reconstructed mesh edges \ensuremath{\mathcal{E}_{mesh}}

\begin{algorithmic}[1]

\STATE Initialize directed edge weights \ensuremath{W} for \ensuremath{\mathcal{G}}

\FOR{each bidirectional edge \ensuremath{(u, v) \in \mathcal{E}}}
    \STATE \ensuremath{w(u, v) \leftarrow \exp(\alpha \cdot (f(v) - f(u)))}
    \STATE \ensuremath{w(v, u) \leftarrow \exp(\alpha \cdot (f(u) - f(v)))}
\ENDFOR

\STATE \textit{// Introduce Super Source for parallel multi-source propagation}
\STATE Add a super source node \ensuremath{v_{super}} to \ensuremath{\mathcal{V}}

\FOR{each seed \ensuremath{s \in \mathcal{S}}}
    \STATE Add directed edge \ensuremath{(v_{super}, s)} with weight \ensuremath{0}
\ENDFOR

\STATE \textit{// Run Single-Source Shortest Path}
\STATE \ensuremath{Dist, Pred \leftarrow \text{Dijkstra}(\mathcal{G}, \text{source}=v_{super})}

\STATE \textit{// Assign Owners via Predecessor Tracking}
\STATE Initialize \ensuremath{Owner(v) \leftarrow -1} for all \ensuremath{v \in \mathcal{V}}

\FOR{each seed \ensuremath{s_i \in \mathcal{S}}}
    \STATE \ensuremath{Owner(s_i) \leftarrow i}
\ENDFOR

\FOR{each valid node \ensuremath{v \in \mathcal{V}} sorted by \ensuremath{Dist(v)}}
    \IF{\ensuremath{Pred(v) \neq v_{super}} \AND \ensuremath{Pred(v) \neq -1}}
        \STATE \ensuremath{Owner(v) \leftarrow Owner(Pred(v))}
    \ENDIF
\ENDFOR

\STATE \textit{// Extract Meeting Frontiers, namely Mesh Edges}
\STATE \ensuremath{\mathcal{E}_{mesh} \leftarrow \emptyset}

\FOR{each edge \ensuremath{(u, v) \in \mathcal{E}}}
    \IF{\ensuremath{Owner(u) \neq Owner(v)} \AND \ensuremath{Owner(u) \neq -1} \AND \ensuremath{Owner(v) \neq -1}}
        \STATE \ensuremath{\mathcal{E}_{mesh} \leftarrow \mathcal{E}_{mesh} \cup \{(Owner(u), Owner(v))\}}
    \ENDIF
\ENDFOR

\STATE \textbf{return} Unique edges in \ensuremath{\mathcal{E}_{mesh}}

\end{algorithmic}
\end{algorithm}

\clearpage

\section{More results about inference-efficiency analysis and other distribution}
\label{appendix:infer}
\textbf{Inference-efficiency.} As shown in Tab. \ref{tab:effi}, \ref{tab:effi2}, our method achieves an average inference time of less than 15 s at 512 resolution, a significant advantage over auto regressive methods; at 1024 resolution, it still completes inference within 1 minute on average.

\textbf{Results on GT mesh distribution.} The shape distribution of the GT mesh (the meshes directly sampled from dataset rather than generated by Trellis.2) is not fully consistent with the shape distribution generated by Trellis.2 from the rendered image as shown in main text, which allows a more thorough comparison of the difference in the operating range between our method and the AR methods. We therefore conducted a new comparison experiment, as shown in Tab. \ref{tab:main-gtdistrib}.

We further evaluate the gain obtained by moving the AR-produced vertices to their nearest points on the GT surface, which can be regarded as a forced restriction of the AR surface solution space. In the table, each value denotes the result after moving the vertices, and the value in parentheses denotes the metric after moving the vertices minus the metric with the original vertices—i.e., the gain that moving the vertices brings to the final result.

The analysis is as follows:

\begin{itemize}
    \item Although moving the AR vertices to the nearest surface does yield some gain, it does not significantly improve the metrics—because the improvement depends not only on placing the vertices on the surface, but also on the actual distribution of the vertices over the surface.
    \item In some cases, we find that part of the AR-generated meshes exhibit defects and produce unreasonable cross-surface triangle connections; these are manifestations of the lack of robustness of the AR paradigm, and all of them degrade the final metrics. 
    \item Whether under the shape distribution of the GT mesh (Tab. \ref{tab:main-gtdistrib}) or under the shape distribution generated by Trellis.2 (Tab. \ref{tab:main}), our method demonstrates a substantial advantage over the AR methods, further proving the robustness and generality of our method.
\end{itemize}

\begin{figure}[t]
    \centering
    \includegraphics[width=1\textwidth]{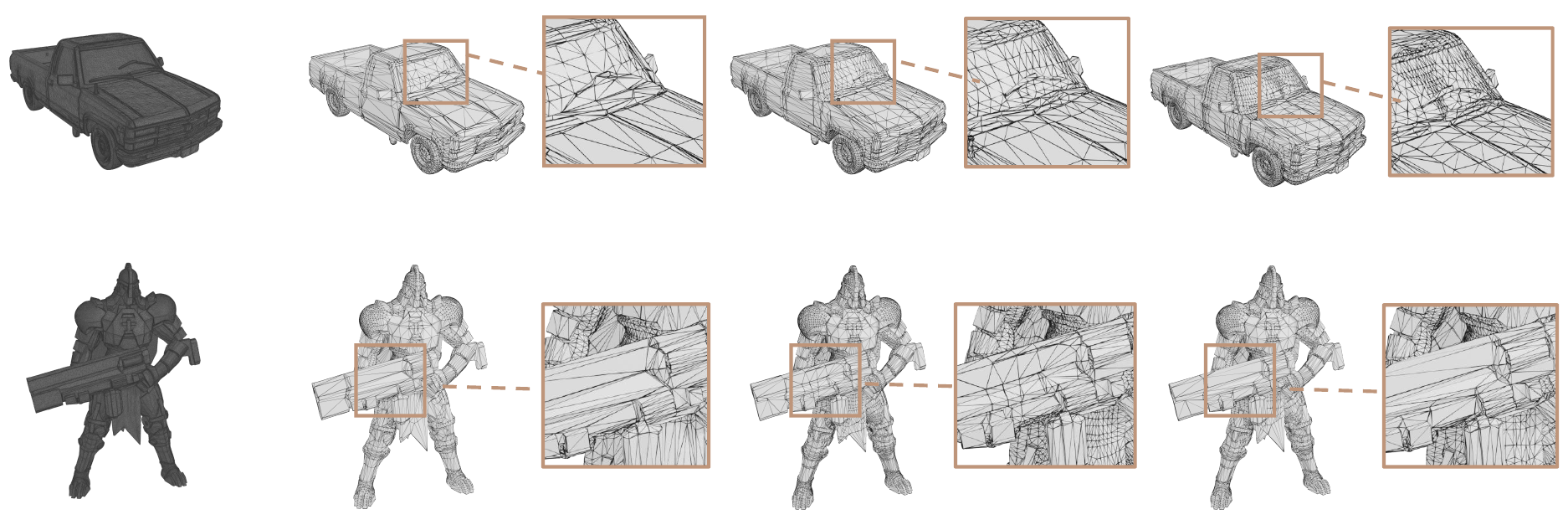}
    \caption{\textbf{Generation diversity.} Changing the sampling noise of BDF latent produces a completely different BDF field and hence a mesh with a different topology.}
    \label{fig:multi}
\end{figure}

\textbf{Generation diversity.} All of our results are meshes reconstructed from the BDF predicted from randomly sampled noise. Changing the sampling noise produces a completely different BDF field and hence a mesh with a different topology. This property is demonstrated in Fig. \ref{fig:multi}.

\begin{table}[t]
\caption{\textbf{Inference time and token limit of AR baselines.}}
\label{tab:effi}
\centering
\begin{tabular}{cccc}
\toprule
Method      & MeshSilksong   & DeepMesh       & MeshRipple     \\ \midrule
Infer Time  & $\sim$ 3.3 min & $\sim$ 8.5 min & $\sim$ 6.5 min \\
Token Limit & 10 k           & 90 k           & 90 k           \\ \bottomrule
\end{tabular}
\end{table}

\begin{table}[]
\caption{\textbf{Inference time of different phases of our method (inference time excluding I/O, e.g., loading image condition, saving files).}}
\label{tab:effi2}
\centering
\begin{tabular}{ccc}
\toprule
Phases           & Infer Time (512) & Infer Time (1024 cascade) \\ \midrule
Sparse Structure & 4 s              & 4.1 s                     \\
Shape Latent     & 2.7 s            & 16.2 s                    \\
Material Latent  & 1.9 s            & 8 s                       \\
BDF Latent       & 2.7 s            & 13.5 s                    \\
Decode Latent    & 1.1 s            & 1.7 s                     \\
Flood Fill       & 1.7 s            & 8.7 s                     \\
Sum              & \textbf{14.1 s}  & \textbf{52.2 s}           \\
Peak Memory      & 7086 MB          & 9841 MB                   \\ \bottomrule
\end{tabular}
\end{table}

\begin{table}[t]
\caption{\textbf{Quantitative Results on the Toys4k Dataset on the GT shape distribution.} For our method, we take the GT low poly meshes of Toys4k as input, and encode them with pretrained Trellis.2 Shape VAE to obtain the shape latent, then we generate BDF latent conditioned on the shape latent. For AR methods, we directly sample point clouds from GT surface as generation condition. }
\label{tab:main-gtdistrib}
\centering

\begin{tabular}{cccccc}
\toprule
\multicolumn{2}{c}{Method}                                                       & MeshSilksong  & DeepMesh      & MeshRipple    & Ours  \\ \midrule
\multirow{3}{*}{\begin{tabular}[c]{@{}c@{}}Toys4k\\ (0 $\sim$ 10 k) \end{tabular}} & CD ($\times e^{-3}$) $\downarrow$   & 2.279(-0.053) & 6.889(-0.017) & 0.814(-0.007) & \textbf{0.089} \\
                                                                          & HD $\downarrow$  & 0.113(-0.002) & 0.155(-0.015) & 0.090(-0.001) & \textbf{0.039} \\
                                                                          & |NC| $\uparrow$& 0.907(+0.011) & 0.914(+0.010) & 0.916(+0.017) & \textbf{0.931} \\ \midrule
\multirow{3}{*}{\begin{tabular}[c]{@{}c@{}}Toys4k\\ (0 $\sim$ 40 k) \end{tabular}} & CD ($\times e^{-3}$) $\downarrow$   & 9.876(-0.100) & 9.863(-0.120) & 1.046(-0.001) & \textbf{0.121} \\
                                                                          & HD $\downarrow$  & 0.223(-0.007) & 0.134(-0.004) & 0.106(-0.001) & \textbf{0.049} \\
                                                                          & |NC|$\uparrow$ & 0.839(+0.024) & 0.853(+0.004) & 0.881(+0.014) & \textbf{0.903} \\ \midrule
\end{tabular}
\end{table}


\newpage
\clearpage
\section*{NeurIPS Paper Checklist}

\begin{enumerate}

\item {\bf Claims}
    \item[] Question: Do the main claims made in the abstract and introduction accurately reflect the paper's contributions and scope?
    \item[] Answer: \answerYes{} 
    \item[] Justification: The main claims in the abstract and introduction accurately summarize the paper's contributions and scope. They are supported by the methodological development and empirical results presented in the main sections of the paper.

\item {\bf Limitations}
    \item[] Question: Does the paper discuss the limitations of the work performed by the authors?
    \item[] Answer: \answerYes{} 
    \item[] Justification: Yes, In the last section and appendix.

\item {\bf Theory assumptions and proofs}
    \item[] Question: For each theoretical result, does the paper provide the full set of assumptions and a complete (and correct) proof?
    \item[] Answer: \answerYes{} 
    \item[] Justification:  We  provide the full set of assumptions and  complete (and correct) proofs.

    \item {\bf Experimental result reproducibility}
    \item[] Question: Does the paper fully disclose all the information needed to reproduce the main experimental results of the paper to the extent that it affects the main claims and/or conclusions of the paper (regardless of whether the code and data are provided or not)?
    \item[] Answer: \answerYes{} 
    \item[] Justification: We have fully detailed the dataset splits, model architectures, training hyperparameters, and evaluation metrics in the "Datasets and Implementation Details" section. We also provided the pseudocode for our core flood-fill algorithm.

\item {\bf Open access to data and code}
    \item[] Question: Does the paper provide open access to the data and code, with sufficient instructions to faithfully reproduce the main experimental results, as described in supplemental material?
    \item[] Answer: \answerNo{}  
    \item[] Justification:  The project page provides a code link, while pretrained checkpoints and complete reproduction instructions will be public soon.

\item {\bf Experimental setting/details}
    \item[] Question: Does the paper specify all the training and test details (e.g., data splits, hyperparameters, how they were chosen, type of optimizer) necessary to understand the results?
    \item[] Answer: \answerYes{} 
    \item[] Justification: All necessary training and testing details, including the choice of optimizer (AdamW), learning rates, batch sizes, and dataset configurations, are explicitly specified in the Experiments section.

\item {\bf Experiment statistical significance}
    \item[] Question: Does the paper report error bars suitably and correctly defined or other appropriate information about the statistical significance of the experiments?
    \item[] Answer: \answerNo{} 
    \item[] Justification: Error bars and statistical significance tests are not reported due to the prohibitively high computational cost and time required to train the 3D diffusion models multiple times from scratch.

\item {\bf Experiments compute resources}
    \item[] Question: For each experiment, does the paper provide sufficient information on the computer resources (type of compute workers, memory, time of execution) needed to reproduce the experiments?
    \item[] Answer: \answerYes{} 
    \item[] Justification: We explicitly reported the compute resources in the Experiments section, including the GPU type (NVIDIA H200), the number of GPUs used (8), and the total training time for both VAE and Flow models.
    
\item {\bf Code of ethics}
    \item[] Question: Does the research conducted in the paper conform, in every respect, with the NeurIPS Code of Ethics \url{https://neurips.cc/public/EthicsGuidelines}?
    \item[] Answer: \answerYes{} 
    \item[] Justification: The research conducted in this paper conforms in every respect with the NeurIPS Code of Ethics.

\item {\bf Broader impacts}
    \item[] Question: Does the paper discuss both potential positive societal impacts and negative societal impacts of the work performed?
    \item[] Answer: \answerYes{} 
    \item[] Justification: We discussed these in Introduction section.
    
\item {\bf Safeguards}
    \item[] Question: Does the paper describe safeguards that have been put in place for responsible release of data or models that have a high risk for misuse (e.g., pre-trained language models, image generators, or scraped datasets)?
    \item[] Answer: \answerNA{} 
    \item[] Justification: The proposed 3D native mesh generation model focuses on foundational geometry synthesis and does not pose a high risk for severe malicious misuse; thus, specific safeguards are not applicable.

\item {\bf Licenses for existing assets}
    \item[] Question: Are the creators or original owners of assets (e.g., code, data, models), used in the paper, properly credited and are the license and terms of use explicitly mentioned and properly respected?
    \item[] Answer: \answerYes{}
    \item[] Justification: We properly cited the original papers and respected the terms of use for the datasets (Texverse, Toys4k) and the backbone pipeline (Trellis.2) used in our experiments

\item {\bf New assets}
    \item[] Question: Are new assets introduced in the paper well documented and is the documentation provided alongside the assets?
    \item[] Answer: \answerNA{} 
    \item[] Justification: This paper focuses on a novel algorithmic representation and generation pipeline, and does not introduce or release any new datasets or similar new assets.

\item {\bf Crowdsourcing and research with human subjects}
    \item[] Question: For crowdsourcing experiments and research with human subjects, does the paper include the full text of instructions given to participants and screenshots, if applicable, as well as details about compensation (if any)? 
    \item[] Answer: \answerNA{} 
    \item[] Justification: This research focuses purely on 3D generative models and does not involve any crowdsourcing or human subjects.

\item {\bf Institutional review board (IRB) approvals or equivalent for research with human subjects}
    \item[] Question: Does the paper describe potential risks incurred by study participants, whether such risks were disclosed to the subjects, and whether Institutional Review Board (IRB) approvals (or an equivalent approval/review based on the requirements of your country or institution) were obtained?
    \item[] Answer: \answerNA{} 
    \item[] Justification:  This research does not involve human subjects, so IRB approval is not applicable.

\item {\bf Declaration of LLM usage}
    \item[] Question: Does the paper describe the usage of LLMs if it is an important, original, or non-standard component of the core methods in this research? Note that if the LLM is used only for writing, editing, or formatting purposes and does \emph{not} impact the core methodology, scientific rigor, or originality of the research, declaration is not required.
    \item[] Answer: \answerNA{} 
    \item[] Justification: Large Language Models (LLMs) were not used as an important, original, or non-standard component of the core methodology in this research.

\end{enumerate}

\end{document}